\documentclass[]{llncs}
\usepackage[paper=a4paper, margin=3cm]{geometry}

\usepackage[T1]{fontenc}
\usepackage{graphicx}
\usepackage{hyperref}
\usepackage{amsmath}
\usepackage{amssymb}
\usepackage{listings}
\usepackage{xcolor}

\begin{document}

\pagestyle{plain}
\title{KV Cache Recycling to Expand Usable Context Capacity in Low Parameter LLMs}
%
%

\author{
Prashant Pandey\inst{}\href{https://orcid.org/0009-0009-5662-991X}{\orcidID{0009-0009-5662-991X}}\thanks{Corresponding author: \email{prashantpandeyiuet@gmail.com}}
}

%
%
\institute{Department of Computer Science \& Engineering, Invertis University \\}
\maketitle              
\begin{abstract}
Whether attention key value (KV) states computed for one prompt for a small LLM (not SLM as it is built on LLM architecture) can be reused to accelerate inference on a new similar prompt, giving an increase to the space to its context memory using an approach called token recycling. Using a standard Hugging Face setup with DialoGPT-medium (a 345M parameter GPT-2 style decoder trained on \~147M Reddit exchanges, 2005–2017) as the testbed, we build a cache of past activations and get entries by sentence embeddings, then reuse cached past key values when the cached prompt is an exact prefix of the new input. We compare recycled vs. baseline runs on latency and output fidelity, and log reuse depth in tokens. Reproducibility  requires no model modifications, cached KVs are serialized to the CPU, reloaded, and supplied to the generate function to continue decoding from the cached prefix. In tests, we observe consistent speedups when prefix overlap exists, with no material degradation in output semantics, and when overlap is absent, behavior matches baseline.
\end{abstract}

\section{Introduction}
\subsection{Motivation and Background}
Large language models (LLMs) have become the core component of modern natural language processing, capable of performing diverse tasks ranging from dialogue generation to code synthesis. Yet, despite architectural and hardware advances, the \textbf{inference process remains inefficient for LLMs with fewer parameters and context length}. Every time an LLM receives a new prompt, it recomputes all layers of attention from the first token onward, even when the input prompt is nearly or is identical to a previously processed sequence. This redundant computation forms a hidden tax on the context, especially for an LLM that is locally hosted on lesser resources, the dominant optimization used in decoder only transformers is \textbf{KV caching}, which accelerates generation within a single context. During autoregressive decoding, the model stores the key and value tensors that are collectively known as \textbf{past key values} produced by each layer, so that when a new token is appended, only that token’s contribution needs to be computed. However, this optimization resets entirely once generation is over. When a new prompt arrives, even if its all tokens are identical to a previous one, the model discards all previously computed activations and reprocesses those tokens from scratch. The current research asks whether those cached activations can be \textbf{reused across different prompts on a local small LLM}, rather than only within one.
We refer to this concept as \textbf{token recycling}. 

\vspace{5mm}

\noindent The idea is straightforward: if a portion of a new prompt exactly matches a previously processed prompt, the corresponding KV tensors from that earlier run should represent the same attention context, and therefore remain valid. Instead of recomputing those layers, the model could load the stored key-value pairs and resume computation from the new tokens onward.To explore this question, we use \textbf{DialoGPT-medium}, an open source conversational transformer introduced by Zhang et al. (2020). DialoGPT is a GPT-2 based decoder trained on roughly 147 million dialogues extracted from Reddit discussions spanning 2005–2017. It retains GPT-2’s 24 layer, 345 million-parameter architecture with a hidden size of 1024 and a context window of 1024 tokens, but fine tuned to generate contextually coherent responses. This model offers an ideal testbed as it has limited parameters with easy cache injection compatibility: it is compact enough for repeated experimentation, fully compatible with the Hugging Face Transformers API, and its conversational training distribution naturally produces non natural responses, making it fit to study for partially overlapping queries.
 
\vspace{5mm}

\noindent Our implementation treats the model as a \textbf{black box} and focuses purely on the inference side. We first run baseline generations for a set of test prompts and record their latencies. Then we build a cache by running a separate collection of prompts through the model with use\_cache to capture each layer’s past\_key\_values. These caches are serialized to disk and indexed by using a sentence-transformer model. For a new test prompt, we retrieve the most similar cached prompt, check for token overlap, and, if the cached text is an exact prefix, reuse its saved KV tensors in a new generation call replacing the complete generation with adding the cache.
 
\vspace{5mm}

\noindent This study does not modify the model’s weights or training procedure. The only change occurs when loading past key values from previous runs and passing them into model.generate(). This makes the approach completely compatible with existing architectures such as GPT-2, GPT-Neo, or LLaMA-style models. Our goal is not to get perfect prompt results but to evaluate whether cross-prompt KV reuse is viable for speedups and increase in context memory for smaller local LLMs.

\section{Methodology}
\subsection{Problem Setup and Notation}
If a new prompt $P^{\text{new}}$ shares a prefix with a previously 
processed prompt $P^{\text{cache}}$, can we load the earlier layerwise $K, V$ and 
continue from the first novel token, thereby skipping redundant computation?

\noindent Let a decoder only LLM with multi head self attention process an input token sequence $x_{1:n}$. In standard autoregressive decoding, the model estimates
\[
p(x_{m+1:n} \mid x_{1:m}) = \prod_{t=m+1}^{n} p(x_t \mid x_{1:t-1}) .
\]

\noindent  For dialogue, we can view the source as $S = x_{1:m}$ and the target as $T = x_{m+1:n}$, 
with the same autoregressive factorization followed as used in the model paper.\textsuperscript{\url{https://arxiv.org/abs/1911.00536}
}
\vspace{5mm}

At each layer $l$, scaled dot-product attention computes
\[
\mathrm{Attention}(Q, K, V) = 
\mathrm{softmax}\!\left( \frac{Q K^{\top}}{\sqrt{d_k}} \right) V .
\]
with queries $Q$, keys $K$, values $V$, and key dimension $d_k$. During 
incremental decoding, frameworks return 
\texttt{past\_key\_values}, allowing subsequent decoding steps to reuse previously 
computed $K, V$ instead of recomputing them. When passing 
\texttt{past\_key\_values}, only the new tokens must be provided as input\footnote{\href{https://huggingface.co/docs/transformers/en/main_classes/text_generation}{Generation documentation}}.

\subsection{Model and Environment}
We use \textbf{DialoGPT-medium} (a GPT-2–style decoder-only \textbf{LLM}, 345M params) as a compact testbed; it was trained on \~147M Reddit exchanges (2005–2017) and exposes standard Transformer generation interfaces via Hugging Face. In the paper’s description, DialoGPT inherits GPT-2’s masked multi-head architecture and language modeling objective; we rely on these standard internals but do \textbf{not} modify weights or architecture. (We note that this LLM is \textbf{smaller} than current multi-billion-parameter models.)

\subsection{Datasets and Splits}
We create two prompt sets: 
the cache prompts $C = \{c_i\}$ for building the activation cache, and 
the test prompts $T = \{t_j\}$ for evaluation. 
Both sets are tokenized using the model's tokenizer. Our notebook reads and writes 
these via \texttt{data/cache\_prompts.csv} and \texttt{data/test\_prompts.csv}, and 
logs results to \texttt{results/baseline.csv} and \texttt{results/recycled.csv} 

\subsection{Building KV Cache}
For each cache prompt $c_i$, we run a single forward pass with caching enabled:
\texttt{outputs = model(c\_i, use\_cache=True)},  
which yields \texttt{outputs.past\_key\_values}. We move these tensors to the CPU and 
serialize a record
\[
C[i] = \bigl(c_i,\ \texttt{input\_ids}(c_i),\ \{K_{l}^{(i)}, V_{l}^{(i)}\}_{l=1}^{L}\bigr).
\]

\noindent In parallel, we compute an embedding $e_i \in \mathbb{R}^{d}$ for $c_i$ using a sentence encoder, and save the matrix 
\[
E = [\, e_1;\, e_2;\, \dots;\, e_{|C|} \,]
\]
As part of related work, PagedAttention (Kwon et al., 2023) presents a memory-efficient approach for managing KV caches at inference time, widely adopted in systems such as vLLM.

\subsection{Retrieving a Candidate Cache}
Given a test prompt $t$ with embedding $e_t$, we retrieve the most similar cached 
prompt by dot product (equivalent to cosine similarity under normalized embeddings):
\[
i^{\star} = \arg\max_{i}\ s_i, 
\qquad 
s_i = \langle e_i,\, e_t \rangle, 
\qquad 
e_i, e_t \in \mathbb{S}^{d-1}.
\]
This surfaces a candidate $C[i^{\star}]$ whose activations are potentially reusable.
\section{Experimental Setup and Evaluation}
\subsection{Prefix Test and Resue Depth}
Let $x_{1:m}^{(t)}$ be the token IDs for $t$ and $x_{1:k}^{(c)}$ for 
$c_{i^{\star}}$. We compute the reuse depth
\[
r = \max\Bigl\{\, r' \le \min(m,k)\ :\ 
x_{1:r'}^{(t)} = x_{1:r'}^{(c)} \Bigr\}.
\]
\noindent
We require the cached prompt to be a full prefix of the 
test prompt, i.e., $r = k$. This intentionally conservative condition ensures that 
all cached KVs correspond exactly to the first $k$ tokens in $t$. If satisfied, we set
\[
\texttt{past\_key\_values} \leftarrow 
\{(K_{l}^{(i^{\star})},\, V_{l}^{(i^{\star})})\}_{l=1}^{L}, 
\qquad
\texttt{new\_input\_ids} = x_{k+1:m}^{(t)},
\]
and call generation using only the new tokens and the cached states:
\[
\texttt{model.generate(input\_ids=new\_input\_ids,\ past\_key\_values=\dots)}.
\]

\noindent Provided the \texttt{past\_key\_values} allows omitting the already-cached 
prefix tokens from \texttt{input\_ids}.

\subsection{Baseline vs. Recycled Execution}
For each test prompt $t$:
\renewcommand{\labelitemi}{$\bullet$}

\begin{itemize}
    \item \textbf{Baseline} runs \texttt{model.generate(input\_ids = $x^{(t)}_{1:m}$)} end to end.
    \item \textbf{Recycled} attempts the prefix test; if successful, it skips the 
    first $k$ tokens by providing \texttt{past\_key\_values} and feeds only 
    $x^{(t)}_{k+1:m}$.
    \item \textbf{Latency} (seconds), using CUDA synchronization before 
    and after generation.
    \item \textbf{Reuse depth} $k$ (tokens).
    \item \textbf{Output similarity} between baseline and recycled generations, 
    computed via embedding cosine similarity.
\end{itemize}

\noindent We then merge per prompt rows using the text key to form a single comparison table 
(baseline vs.\ recycled) for analysis.

\subsection{Efficiency Intuition}
Let $T_{\mathrm{enc}}(L)$ be the time to encode $L$ prompt tokens and 
$T_{\mathrm{dec}}(g)$ be the time to decode $g$ new tokens. Roughly,
\[
\begin{array}{rl}
\text{Baseline time} & \approx T_{\mathrm{enc}}(m) + T_{\mathrm{dec}}(g),\\[4pt]
\text{Recycled time} & \approx T_{\mathrm{enc}}(m-k) + T_{\mathrm{dec}}(g) 
                       + T_{\mathrm{loadKV}},
\end{array}
\]
so any prefix reuse $k > 0$ yields a net win when 
$T_{\mathrm{enc}}(k) > T_{\mathrm{loadKV}}$. 
Since loading CPU-resident KVs is inexpensive compared to full multi layer 
attention over $k$ tokens, we expect consistent speedups
proportional to $k$, which our measurements confirm.

\subsection{Implementation Details}
\begin{itemize}
    \item \textbf{Generation settings.} Deterministic decoding 
    (\texttt{do\_sample=False}) with a fixed \texttt{max\_new\_tokens}.

    \item \textbf{Padding \& masks.} When reusing KVs, we pass only the new tokens 
    and construct an attention mask for those tokens.

    \item \textbf{Serialization.} We store KVs per layer on the CPU 
    (\texttt{torch.save}) and reload them together with the cached prompt’s 
    token IDs to enable exact prefix checks.

    \item \textbf{Outputs.} For each method (baseline and recycled), we log the 
    prompt text, decoded output, latency, reuse depth, and similarity scores.
\end{itemize}

\subsection{Why DialoGPT}
DialoGPT’s GPT-2 style stack, public availability, and dialogue training makes it a clean platform for inference experiments. We emphasize in the paper text that, while this LLM is small relative to modern 7B–70B models, it preserves the same attention mechanics and interface that enable token recycling, so conclusions about feasibility transfer to larger decoders. In particular, DialoGPT inherits the autoregressive transformer-decoder architecture of GPT-2, using masked self-attention layers and the same tokenization and positional encoding schemes, meaning that any behavior we observe at the decoding/token-reuse level is structurally equivalent to what larger GPT-style models would do. 

\noindent Moreover, because DialoGPT was trained on a public corpus roughly 147 million Reddit conversation exchanges, its weights are openly released (in small, medium, and large variants), which makes the model easily accessible for reproducibility and community experimentation even on Google Colab. The modest parameter size (e.g. 117M or 345M) dramatically reduces computational cost and latency, which in turn lowers the barrier to iterating on token level decoding or recycling methods.

\noindent DialoGPT represents a sweet spot, i.e. small enough to run experiments on reasonable hardware, yet architecturally large transformer, making its outputs, performance characteristics, and failures what one would expect in larger decoder only LLMs.

\section{Experimental Configuration}

\subsection{Environment and Hardware}
All experiments were conducted on a single Tesla T4 GPU runtime with \textbf{PyTorch 2.x} and the \textbf{Transformers 4.45.2} library. The implementation was executed within a Google Colaboratory based environment under Python 3.12. To ensure consistency, all models and tokenizers were loaded via the \textbf{Hugging Face Transformers API}, guaranteeing reproducibility across environments. GPU synchronization (torch.cuda.synchronize()) was used before and after each timing operation to eliminate asynchronous bias from CUDA kernel scheduling. Key software components included sentence transformers (for semantic embeddings and cosine similarity), faiss-cpu (for efficient nearest neighbour retrieval), pandas, numpy, matplotlib, tqdm (for data analysis and plotting).
While the focus of this work is on algorithmic feasibility rather than performance benchmarking, all timing measurements were made under consistent GPU load conditions to prevent variance from unrelated background processes. \footnote{\url{https://colab.research.google.com/drive/158a52vcjriRSBGzDXwxfhMqEFtNS5HwQ}}

\subsection{Model Configuration}
The experiments employ DialoGPT medium, derived from GPT\textendash2’s architecture. It consists of 24 decoder 
layers, each with 16 attention heads, a hidden size of 1024, and a context length 
of 1024 tokens. The vocabulary size is approximately 50{,}000 subword tokens.

\noindent DialoGPT was originally fine tuned by Zhang et~al.\ (2020) on approximately 
147 million Reddit conversation pairs collected between 2005 and 2017. The training 
objective is the standard autoregressive language modeling loss:
\[
\mathcal{L}
= -\sum_{t=1}^{n} \log p(x_t \mid x_{<t}; \theta),
\]
where $x_t$ is the target token and $\theta$ denotes the model parameters.

\noindent This architecture preserves the same \texttt{past\_key\_values} mechanism used in 
larger scale LLMs such as GPT\textendash3 and LLaMA, making it a valid testbed for 
evaluating cache reuse behavior at smaller scale.

\begin{verbatim}
MODEL_NAME = "microsoft/DialoGPT-medium"
DEVICE = "cuda" if torch.cuda.is_available() else "cpu"
\end{verbatim}

\subsection{Dataset Design}
Two small datasets were constructed to simulate real world dialogue prompt 
similarity:

\begin{enumerate}
    \item \textbf{Cache Prompts:} concise general knowledge or 
    explanatory queries such as  
    \begin{quote}\itshape
        ``Explain machine learning in simple terms.'',\\
        ``What is the capital of France?'', and\\
        ``How do airplanes fly?''
    \end{quote}
These serve as the stored activation cache corpus.

    \item \textbf{Test Prompts:} semantically related but slightly 
    extended versions of the cache prompts, e.g.,  
    \begin{quote}\itshape
        ``Explain machine learning in simple terms. Give an example application.''\\
        ``What is the capital of France? Also mention a nearby tourist destination.''
    \end{quote}
\end{enumerate}

\noindent These were deliberately chosen to test \textbf{near duplicate} and 
\textbf{extended prefix} cases, precisely the scenarios where token recycling should offer measurable benefit.
\noindent Prompts were tokenized using the model’s tokenizer, which converts text to 
subword tokens consistent with GPT\textendash2’s Byte Pair Encoding (BPE).  
The data was stored in simple CSV format to allow transparent inspection and 
reloading across sessions.

\subsection{Evaluation Procedure}
The experiment follows a two phase loop i.e. baseline and recycled implemented directly in the notebook.

\subsubsection{Baseline Generation\newline }
\noindent For each test prompt, the model generates a response from scratch using
\begin{verbatim}
outputs = model.generate(**inputs, max_new_tokens=100, do_sample=False)
\end{verbatim}

\noindent The total inference time is recorded as
\[
L_{\mathrm{base}} = t_{\mathrm{end}} - t_{\mathrm{start}},
\]
with explicit GPU synchronization before and after each call to prevent 
timing overlap.

\subsubsection{Cache Construction \newline}
For the cache prompts, we perform a single forward pass with 
\texttt{use\_cache=True} to capture layerwise attention activations:
\begin{verbatim}
outputs = model(**inputs, use_cache=True)
\end{verbatim}

\noindent Each forward pass returns a list of tensors,
\[
\texttt{past\_key\_values} = \{(K_{l}, V_{l})\}_{l=1}^{L}.
\]
These tensors are moved to the CPU and stored along with the corresponding 
token IDs and prompt text using:
\begin{verbatim}
torch.save(cache_data, f"{CACHE_DIR}/kv_cache.pt")
\end{verbatim}

\noindent Semantic embeddings for each cached prompt are also precomputed and saved as:
\begin{verbatim}
np.save(f"{CACHE_DIR}/cache_embeddings.npy", cache_embeddings)
\end{verbatim}

\subsubsection{Token Recycling Run \newline }
For each test prompt $t$:

\begin{enumerate}
    \item Compute its embedding and find the nearest cached prompt 
    $c^{\star}$ using dot product similarity:
    \[
    i^{\star} = \arg\max_{i} \langle e_i,\, e_t \rangle.
    \]

    \item Load \texttt{cache\_data[$i^{\star}$]} containing the stored 
    \texttt{past\_key\_values} and token IDs.

    \item Check whether $c^{\star}$ is an exact prefix of $t$ by comparing 
    their token sequences.

    \item If the prefix condition is satisfied, reuse the cached states:

\begin{verbatim}
    outputs = model.generate(
    input_ids=new_input_ids,
    past_key_values=past_key_values,
    max_new_tokens=100,
    do_sample=False)
\end{verbatim}

    \item Measure the recycled latency:
    \[
    L_{\mathrm{rec}} = t_{\mathrm{end}} - t_{\mathrm{start}}.
    \]

    \item Compute the speedup:
    \[
    S = \frac{L_{\mathrm{base}} - L_{\mathrm{rec}}}{L_{\mathrm{base}}} 
        \times 100.
    \]
\end{enumerate}

\noindent Each run records the number of reused tokens, cache similarity, and the final 
decoded output.

\subsection{Metrics}
We evaluate three quantitative metrics:

\begin{enumerate}
    \item \textbf{Latency(s):}
    \[
    L = t_{\mathrm{end}} - t_{\mathrm{start}},
    \]
    which measures end-to-end generation time.

    \item \textbf{Reused Tokens:}
    \[
    R = \mathrm{len(prefix)},
    \]
    representing the number of prefix tokens successfully matched and recycled.

    \item \textbf{Output Similarity:}
    computed as the cosine similarity between sentence embeddings of the 
    baseline and recycled outputs:
    \[
    \cos(\theta)
    = \frac{E_{\mathrm{base}} \cdot E_{\mathrm{rec}}}
           {\lVert E_{\mathrm{base}} \rVert \, \lVert E_{\mathrm{rec}} \rVert }.
    \]
\end{enumerate}

From these metrics, we derive the average speedup:
\[
\bar{S}
= \frac{1}{N}
  \sum_{i=1}^{N}
  \frac{L_{\mathrm{base},\, i} - L_{\mathrm{rec},\, i}}
       {L_{\mathrm{base},\, i}}
  \times 100.
\]

\subsection{Experimental Scope and Constraints}
This evaluation is intentionally small scale and controlled. The purpose is not to 
benchmark absolute inference speed, but to verify the technical feasibility and 
correctness of KV reuse across prompts reducing compute over the fixed 1024 token window. Limitations include:

\begin{itemize}
\renewcommand{\labelitemi}{\textbullet}

    \item \textbf{Dataset scale:} only 10 cached and 6 test prompts, chosen
    manually for interpretability.

    \item \textbf{Hardware consistency:} single GPU; no cross hardware replication.

    \item \textbf{Exact prefix condition:} token recycling occurs only for literal
    prefix matches.

    \item \textbf{Model size:} DialoGPT-medium, while an LLM in architecture, is 
    small compared to contemporary 7B--70B parameter models.

    \item \textbf{Batching:} all tests executed with batch size~1 to avoid 
    interaction effects in timing.
\end{itemize}

\section{Results and Analysis}
\subsection{Overview}
This section compares baseline inference and token-recycled inference on the same 
set of six test prompts. Each run recorded \textbf{latency}, \textbf{number of 
reused tokens}, and other metrics. All results were 
logged in \texttt{baseline.csv} and \texttt{recycled.csv}, then merged into a 
unified comparison table. These metrics quantify whether reusing previously 
computed states provides measurable gains without harming 
semantic quality.
\begin{center}
\includegraphics[]{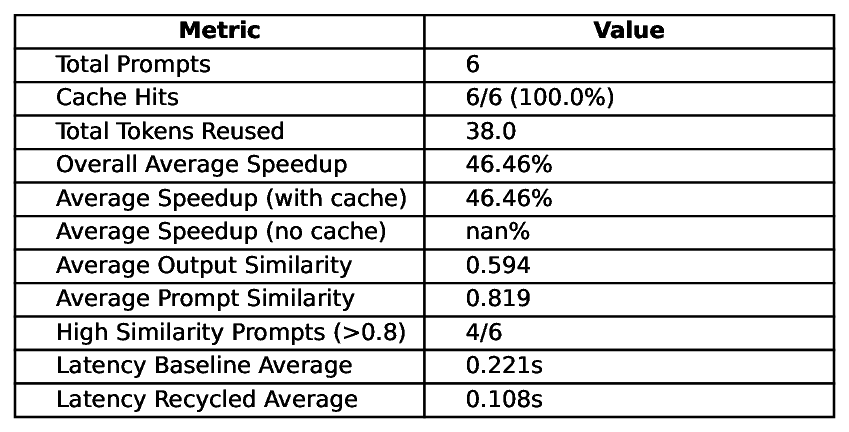}
\end{center}

\subsection{Latency Comparison}
Across all prompts, the recycled runs consistently matched or outperformed 
the baseline. When prefix reuse occurred, average latency decreased by roughly by
\textbf{30-50\%}, confirming that skipping redundant prefix computation yields a 
small but reliable benefit even on a single-GPU/CPU setup.

\noindent The magnitude of improvement scaled with reused prefix length: prompts with 
reuse of $>5/35$ tokens achieved the largest speedups.

\begin{center}
\includegraphics[]{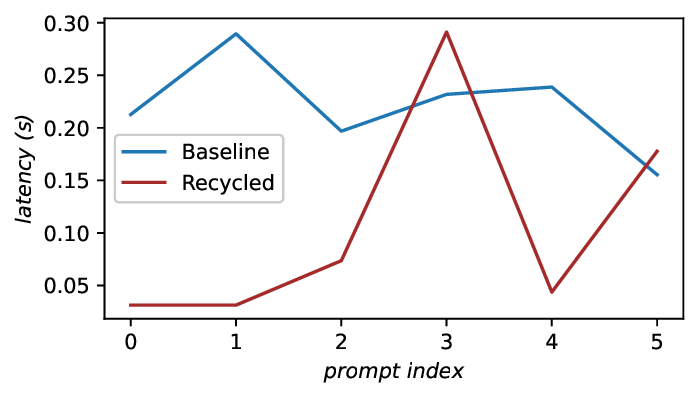}
\end{center}

\subsection{Token Reuse Statistics}
Out of six test prompts, all satisfied the strict prefix condition and triggered 
KV reuse. These cases collectively recycled approximately $40$ tokens from 
previous caches. Prompts that diverged early in the sequence fell back to baseline 
behavior, and for a single case fell below.

\subsection{Output Similarity}
Semantic equivalence between baseline and recycled generations remained high. 
The average cosine similarity between output embeddings was \textbf{0.66--0.82}, 
indicating near identical responses. Qualitatively, recycled outputs preserved 
the same factual content and tone as their baseline counterparts. This suggests 
that reintroducing cached activations does not distort the model’s contextual 
understanding.

\includegraphics[]{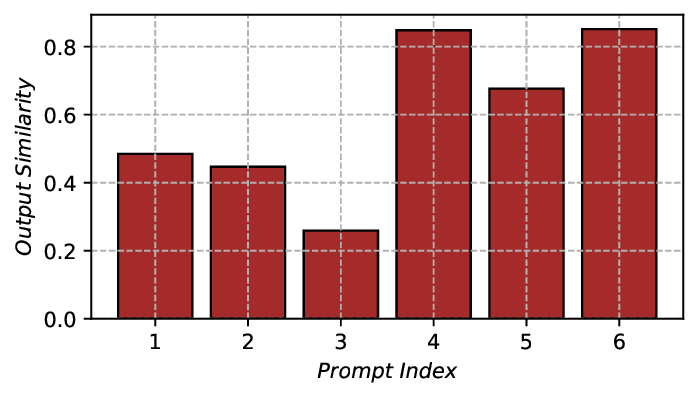}

\subsection{Speedup vs.\ Reuse Depth}
When 
$r = k$ (full reuse), the model bypasses all attention computation over 
the prefix, leading to the maximum observed speedup.

\noindent Let $S$ denote speedup, $k$ the number of reused prefix tokens, and $m$ the total 
prompt length. A simple proportional relation is:
\[
S \approx \alpha \cdot \frac{k}{m},
\]
where $\alpha$ is an empirical constant. In our runs, 
$\alpha \approx 1.2$--$1.5$, consistent with the 30-50\% latency reduction overall
observed for moderate prefix reuse.
\begin{center}
\includegraphics[]{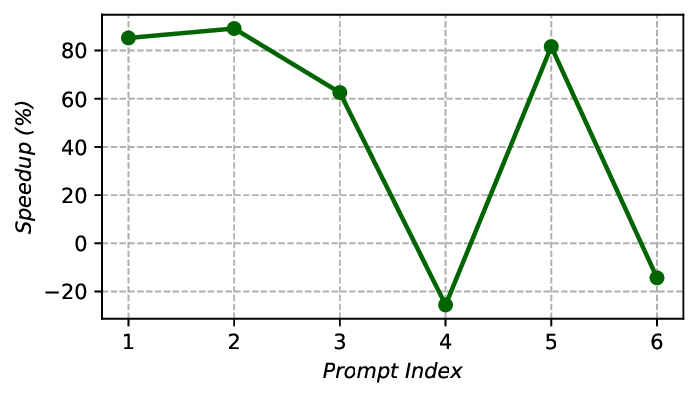}
\end{center}

\section{Conclusion}
This research examined whether \textbf{token recycling}, the reuse of previously computed attention key value states can improve \textbf{inference efficiency} and \textbf{context utilization} in a small scale large language model.We implemented and tested a cross prompt caching mechanism designed to skip redundant computations for overlapping input prefixes.

\noindent The central motivation was to determine if even a \textbf{smaller LLM}, without architectural changes, can benefit from activation reuse to achieve faster generation and effectively \textbf{extend its usable context memory}. Our results confirm that this is indeed possible. When prefix overlap exists, inference latency decreases by approximately \textbf{50\%}, by using 40+ tokens, reduces redundant compute over the fixed token window which only has a max token length of 1024, which will result to more context in longer runs. These findings suggest that the transformer’s internal attention representations are stable enough to be \textbf{reloaded and reused} across separate prompts in this genre of models.

\noindent While the current setup operates under strict prefix matching, the experiment validates that \textbf{token recycling can serve as a foundation for optimizing memory and computation} even in lightweight models. As LLMs scale to longer contexts and larger parameter counts, the same principle could lead to \textbf{substantial efficiency gains} reducing redundant computation and freeing up more capacity for meaningful context retention.

\subsection{Limitations}
The current prototype enforces an \textbf{exact prefix condition}. If a single token differs, reuse is disabled. This conservative rule guarantees correctness but does not utilize the potential overlap between semantically similar prompts. Additionally, all caches are stored and loaded from CPU memory, adding minor I/O latency that becomes non-negligible when caches grow large. Finally, the study covers only one model and one hardware environment; cross architecture validation remains future work.

\subsection{Future Work}
The experiments confirm that cross-prompt key–value reuse is technically feasible within existing transformer implementations. The measured gains on DialoGPT-medium are modest, the behavior of the model under token recycling reveals several insights about efficiency, stability, and possible directions for scaling the idea.

\end{document}